\documentclass{article}
\usepackage{graphicx}
\usepackage{amsmath,amsfonts}
\usepackage{algorithm}
\usepackage{wrapfig}
\usepackage[noend]{algpseudocode}
\usepackage{subcaption}
\usepackage[toc,page]{appendix}
\usepackage[final]{corl_2020} 

\title{Reinforcement Learning with Videos: \\Combining Offline Observations with Interaction}

\newcommand{\REV}[1]{#1}

\newcommand{\obsgen}{\mathbf{s}}
\newcommand{\obsint}{\mathbf{s}_{int}}
\newcommand{\obsobs}{\mathbf{s}_{obs}}

\newcommand{\featgen}{\mathbf{h}}
\newcommand{\featint}{\mathbf{h}_{int}}
\newcommand{\featobs}{\mathbf{h}_{obs}}

\newcommand{\actint}{\mathbf{a}_{int}}
\newcommand{\actobs}{\mathbf{a}_{obs}}

\newcommand{\predactint}{\mathbf{\hat{a}}_{int}}

\newcommand{\rewardint}{\mathbf{r}_{int}}

\newcommand{\predrewardobs}{\mathbf{\hat{r}}_{obs}}

\newcommand{\poolint}{\mathcal{D}_{int}}
\newcommand{\poolobs}{\mathcal{D}_{obs}}

\newcommand{\inv}{f_{inv}}
\newcommand{\invparam}{\theta}
\newcommand{\enc}{f_{enc}}
\newcommand{\encparam}{\psi}

\newcommand{\discrim}{f_{discr}}
\newcommand{\discrimparam}{\phi}

\newcommand{\heading}{-1.75mm}

\author{
    Karl Schmeckpeper$^1$, Oleh Rybkin$^1$, Kostas Daniilidis$^1$, Sergey Levine$^2$, and Chelsea Finn$^3$ \\
    University of Pennsylvania$^1$, University of California, Berkeley$^2$, Stanford University$^3$ \\
    \texttt{karls@seas.upenn.edu}
}

\begin{document}
\maketitle


\begin{abstract}
    Reinforcement learning is a powerful framework for robots to acquire skills from experience, but often requires a substantial amount of online data collection. As a result, it is difficult to collect sufficiently diverse experiences that are needed for robots to generalize broadly. Videos of humans, on the other hand, are a readily available source of broad and interesting experiences. In this paper, we consider the question: can we perform reinforcement learning directly on experience collected by humans? This problem is particularly difficult, as such videos are not annotated with actions and exhibit substantial visual domain shift relative to the robot's embodiment. To address these challenges, we propose a framework for reinforcement learning with videos (RLV).
    RLV learns a policy and value function using experience collected by humans in combination with data collected by robots. In our experiments, we find that RLV is able to leverage such videos to learn challenging vision-based skills with less than half as many samples as RL methods that learn from scratch.\footnote{This paper has been updated from the published version.  A more detailed description of the changes is available in Appendix~\ref{sec:changes}.}
\end{abstract}


\begin{wrapfigure}{R}{0.45\textwidth}
\vspace{-8mm}
    \centering
    \includegraphics[width=0.85\linewidth]{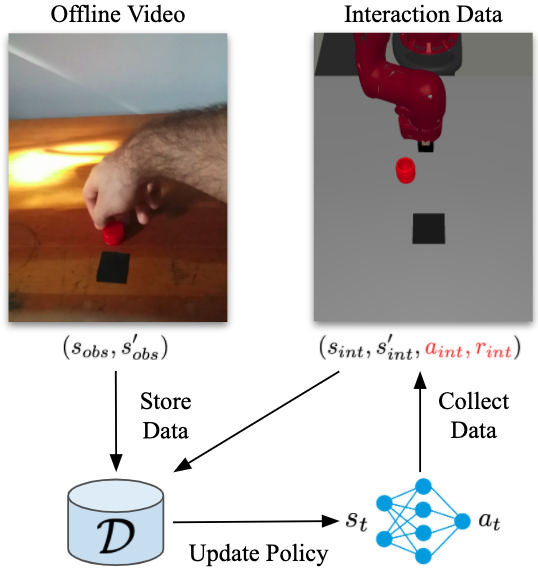}
    \caption{\small \textbf{Reinforcement learning with videos}.
    We study the setting where 
    observational data is available, in the form of videos (top left). Our method can leverage such data to improve reinforcement learning by adding the videos to the replay buffer and directly performing RL on the observational data,
    while overcoming the challenges of unknown actions and domain shift between observation and interaction data.}
    \vspace{-0.25in}
    \label{fig:teaser}
\end{wrapfigure}

\vspace{\heading}
\section{Introduction}
\vspace{\heading}
\label{sec:introduction}

Reinforcement learning is a powerful tool for robots to automatically acquire behaviors, but requires a significant amount of online trial-and-error data collection. While one solution is to build more data-efficient algorithms, advances in other areas of deep learning \cite{krizhevsky2012imagenet,devlin2018bert}
suggest that \textit{large} and \textit{diverse} real-world datasets are critical for broad generalization and high performance. Therefore, we instead look
to videos of humans as a readily available source of broad and interesting training data. We propose to use videos of people within a reinforcement learning framework to augment the data acquired by a robot. If this were feasible, it would represent an important first step towards being able to tap into cheap and readily available datasets of diverse human behaviors when training robots in the real world.
Performing reinforcement learning directly from videos of human-provided behaviors
presents multiple challenges. First, the robot must be able to update its policy using observations without any corresponding actions or rewards. Second, the algorithm must be able to account for domain shift from differences in the action space, morphology, viewpoint, and environment, as humans look different and have different degrees of freedom than most robotic manipulators.
While prior works have made progress on imitation learning of action-free demonstrations~\cite{torabi2018behavioral,edwards2018imitating}, we specifically focus on reinforcement learning since widely-available video data may not perform the task optimally, may perform tasks in a way that is suboptimal for robot morphologies, or may contain trajectories of many distinct tasks.
Even perfect imitation may be unable to learn successful policies from such data, but reinforcement learning, which can learn from both successful and unsuccessful trials, should be able to learn successful policies while better leveraging the information available in the observation data.


Unlike policies learned via imitation learning, value functions acquired with reinforcement learning can capture general-purpose information. 
For example, a door rotates around its hinges regardless of whether its handle was pulled by a person or a robot.
Videos of humans contain insight into object interactions, semantically meaningful tasks, and the physics of the environment.
We therefore aim to leverage such observational data in a reinforcement learning framework, and overcome the aforementioned challenges through a simple approach that infers actions and rewards for the observation data from domain-invariant representations of the observed images.

The main contribution of this paper is a framework for reinforcement learning from videos (RLV) that leverages both offline observation data and online interaction.
We instantiate this framework 
and demonstrate that this approach allows a robot to learn from observational robot or human demonstrations, significantly improving the speed of training.
In a series of simulation experiments, including challenging vision-based robotic manipulation tasks, we find that RLV can leverage observational demonstrations of varying quality to learn tasks with \emph{half} the number of trials, compared to standard RL and prior works on imitation from observation. Further, we find that
RLV can extend to real videos of humans such as those shown in Figure~\ref{fig:teaser}.



\vspace{\heading}
\section{Related Work}
\vspace{\heading}
\label{sec:related_work}



\textbf{Learning from demonstration.}
Early work on learning from demonstration focused on high-quality demonstrations and tried to match the policy of the expert \cite{pomerleau1991efficient,billard2008survey,argall2009survey,ziebart2008maximum}. 
There have been many recent works that study various facets of imitation learning
\cite{finn2016guided,ho2016generative,finn2016connection,fu2017learning,sharma2018multiple,zhang2018deep,rahmatizadeh2018vision,florence2019self,brown2019extrapolating}. 
However, these works assume access to the demonstrator's actions, which often requires extra instrumentation or might even be impossible if morphologies of the demonstrator and the agent are different. In contrast, our method does not require the observation data to contain actions.



\textbf{Imitation learning without actions.} 
In situations such as learning from human demonstrations, only observations of the demonstrator are available, while the actions are not.  
Several approaches to learning only from observations exist, such as matching the demonstrated state transitions \cite{stadie2017third,torabi2018behavioral,torabi2018generative,lee2019efficient,edwards2018imitating,shaoconcept2robot} or training a reward function to encourage behavior similar to that in the demonstrations \cite{sermanet2016unsupervised,sermanet2018time,liu2018imitation,torabi2019imitation,smith2019avid,yu2018one,aytar2018playing,bonardi2020learning}. 
However, these methods are all limited by the performance of the demonstrator that they try to imitate. In contrast, our method integrates observation data into a reinforcement learning pipeline and can improve over the observation data.







\textbf{Reinforcement learning with demonstrations.}
Recent work has used demonstration trajectories to improve performance of reinforcement learning agents \cite{vecerik2017leveraging,rajeswaran2017learning,nair2018overcoming,reddy2019sqil,peng2019advantage,nair2020accelerating}. In contrast to standard imitation approaches, these methods allow improving over the expert performance as the policy can be further fine-tuned via reinforcement learning. We use the same principle to design RLV, however, unlike these works, we don't assume that the demonstration data includes actions.

More closely related to this work, recent approaches have pushed toward removing assumptions on the demonstration data.
\citet{edwards2020estimating} propose learning from offline data without actions by training a state-next state value function and a forward dynamics model. \citet{schmeckpeper2019learning} train a model-based agent that can incorporate data without actions and with domain shift, such as learning robotic tasks from videos of humans. In contrast to these two works, RLV does not require that a good dynamics model be learned, and handles more complex domain shift with different agent morphologies, viewpoints, and background. 
Concurrent work by \citet{chang2020semantic} leverages offline videos for navigation tasks, but not address the challenge of differences in morphology and the action space between the offline observations and the robot data, which is required to learn object manipulation with human videos. By handling more complex types of domain shift, our approach takes a step toward learning visual manipulation tasks with diverse open-world human data.

\textbf{Domain adaptation.}
Data collected from different agents can be very dissimilar.  To overcome the domain shift between different agents, one group of methods used unpaired image-to-image translation \cite{taigman2016unsupervised, zhu2017unpaired, bousmalis2017unsupervised, hoffman2018cycada} to transform samples in a source domain into samples from the target domain. This approach has been used for sim-to-real adaptation \cite{zhang2019vr,rao2020rl} and for learning from demonstrations \cite{liu2018imitation,sharma2019third,smith2019avid}.
A second group of methods, based on domain confusion \cite{tzeng2014deep,ganin2015unsupervised,ganin2016domain,zhuang2017supervised,tzeng2017adversarial}, have sought to learn a set of domain-invariant features that still contain all the required information to complete the task \cite{stadie2017third,tzeng2015adapting,roy2020visual,choi2020cross,kimdomain}.
In contrast to these works, we leverage adversarial domain confusion for the task of reinforcement learning with offline observations.

\vspace{\heading}
\section{Reinforcement Learning with Videos}
\vspace{\heading}

\begin{figure}[t]
\vspace{-4mm}
    \centering
    \includegraphics[width=1\linewidth]{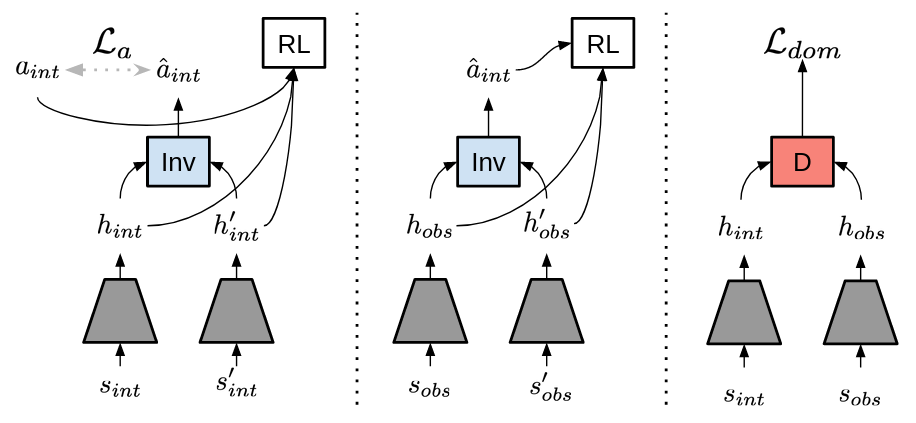}
\vspace{-5mm}
    \caption{\small Components of reinforcement learning with offline videos.
    \textbf{Left:} a batch with samples ($\obsint, \actint, \obsint', \rewardint$) is sampled from the action-conditioned replay pool, $\poolint$, and the observations are encoded into features $\featint, \featint'$. An inverse model is trained to predict the action $\actint$ from the features $\featint, \featint'$. \textbf{Middle:} the inverse model is used to predict the missing actions in the offline videos, $\predactint$, in the robot's action space, from features $(\featobs, \featobs')$ that were extracted from observations $(\obsobs, \obsobs')$. To obtain the missing rewards $\predrewardobs$, we label the final step in the trajectory with a large reward and other steps with a small reward.
    \textbf{Right:} we use adversarial domain confusion to align the features from the action-conditioned data, $\featint$ with the features from the action-free data, $\featobs$. 
    Finally, we use an off-policy reinforcement learning algorithm on the resulting batch $(\left (\featint, \featobs), (\actint, \predactint), (\featint', \featobs'), (\rewardint, \predrewardobs) \right)$. By overcoming the challenges of missing actions, rewards, and the presence of domain shift, we are able to effectively use the observation data to improve performance of a reinforcement learning agent.
    }
    \label{fig:model_architecture}
\vspace{-5mm}
\end{figure}

To learn from observations of another agent, several challenges must be overcome.
The observational data lacks actions and rewards, and may exhibit substantial domain shift relative to the robot.
In this section, we describe our framework for learning with observational data as well as the specific techniques we use for estimating the rewards, estimating the actions, and handling domain shift.


\vspace{\heading}
\subsection{Problem Formulation}
\label{sec:preliminaries}
\vspace{\heading}

We formulate the problem as a Markov decision process (MDP), defined by the tuple $(\mathcal{S}_{int}, \mathcal{A}_{int}, P, R)$ where $\obsint \in \mathcal{S}_{int}$ is the state space, $\actint \in \mathcal{A}_{int}$ is the action space, $P(\obsint' | \actint, \obsint)$ is the environment's dynamics, and $R(\obsint, \actint)$ is the reward function.
The agent is initially provided with a set of observations $\{(\obsobs, \obsobs')_{1:t} \}$ from another agent such as a human, that are modeled as coming from another MDP, $(\mathcal{S}_{obs}, \mathcal{A}_{obs}, P, R)$, which has a distinct state and action space, but whose dynamics and reward function are isomorphic to the original MDP, meaning that there exists some transformation of one state and action space into the other that preserves dynamics and rewards. Formally, this may be written as the assumption that $P(\obsint' | \actint, \obsint) = P(g_o(\obsobs') | g_a(\actobs), g_o(\obsobs))$, where $g_o$ and $g_a$ are unknown invertible functions.
While this assumption represents a simplification, in that the human's state is not actually isomorphic to that of a robot, it does allow us to bridge the gap between the two MDPs with simple domain adaptation methods, as we discuss in Section~\ref{sec:domain_adaptation}, because the isomorphism property implies that there exists a third representation that is \emph{invariant} with respect to the two MDPs.
We further assume that the agent that collected the observational data was executing a successful, but potentially sub-optimal policy.

The robot
may collect additional experience by executing actions, $\actint \in \mathcal{A}_{int}$, to interact with its environment, allowing it to collect observations, $\obsint \in \mathcal{S}_{obs}$, and rewards, $\rewardint \in R_{int}$
The agent seeks to learn a policy that maximizes the expected return in the current environment.

\vspace{\heading}
\subsection{Overview}
\vspace{\heading}



An overview of our method is in Figure~\ref{fig:model_architecture}.
To learn from both offline observations and online interaction, we maintain two replay pools, one containing the action-free observation data, $\left( \obsobs, \obsobs' \right) \in \poolobs$ and one containing the action-conditioned interaction data, $\left( \obsint, \actint, \obsint', \rewardint \right) \in \poolint$.
The pool of interaction data, $\poolint$, is updated during training, while the pool of observation data, $\poolobs$, only contains the initial set of observations.

To overcome domain shift, we learn to embed each observation $\obsgen$ to a domain-invariant feature vector $\featgen$.
This is discussed in Section~\ref{sec:domain_adaptation}.
In order to include the encoded observation tuples, $(\featobs, \featobs')$, in the RL process, they must be annotated with actions and rewards. Drawing on the isomorphism assumption discussed in Section~\ref{sec:preliminaries}, we propose to learn an inverse model $\inv$ that can map the invariant feature tuples $(\featgen,\featgen')$ to robot actions $\actint$. Since the features are invariant between human and robot observations, we can train this inverse model on the robot data, and then use it to annotate human data, as we will discuss in Section~\ref{sec:action_prediction}.
We also generate rewards for the observation data using a simple scheme described in Section~\ref{sec:reward_generation}.
All sampled tuples from \emph{both} replay pools are then combined into a batch and passed into the reinforcement learning algorithm, thereby allowing our method to perform reinforcement learning directly on both the interaction data and the observation data.
This approach is detailed in Algorithm~\ref{alg:rl}.

\begin{algorithm}
\caption{Reinforcement Learning with Videos (RLV)}
\label{alg:rl}
\begin{algorithmic}[1]
\footnotesize
\State Initialize replay buffer $\poolint \gets \{\}$ 
\Comment{Initialize replay pools and networks}
\State Fill buffer $\poolobs \gets \{(\obsobs, \obsobs')_{1:t} \}$ with observed data
\State RL.init(), $\enc$.init(), $\inv$.init()
\For{each iteration}
    \For{each environment step}
    
        \State Sample transition $\left(\obsint, \actint, \obsint', \rewardint\right)$ using RL's exploration policy
        \State $\poolint \gets \poolint \cup \left\{ \left( \obsint, \actint, \obsint', \rewardint \right) \right\}$
    \EndFor
    
    \For{each gradient step}
    
        \State $\{ (\obsint, \actint, \obsint', \rewardint)_{1:n}\} \sim \poolint$ \Comment{Sample from replay pools}
        \State $\{(\obsobs, \obsobs')_{1:m}\} \sim \poolobs$

        \State $\featint \gets \enc(\obsint; \encparam)$
        \Comment{Extract feature representations}
        \State $\featobs \gets \enc(\obsobs; \encparam)$
        
        
        \State $\predactint = \inv(\featobs, \featobs';\invparam)$ \label{alg:line:actions}
        \Comment{Generate actions \& rewards for observation data}

        \State $\predrewardobs = R(\obsobs, \obsobs')$ with $R$ defined in Equation~\ref{eqn:reward_generation}

        \State RL.train\_step$\left( \left\{ (\featint, \actint, \featint', \rewardint)_{1:n}\} \cup \{(\featobs, \predactint, \featobs', \predrewardobs)_{1:m} \right\} \right)$
        \Comment{Train RL}
        
        \State Update $\encparam$, $\invparam$, $\discrimparam$ with joint optimization objective (Equation~\ref{eqn:combined_loss})
    \EndFor
\EndFor
\end{algorithmic}
\end{algorithm}
This framework can be combined with any method of estimating the actions and the rewards of the observational data.
We next present the simple approaches that we use for estimating the action and for estimating the reward.

\vspace{\heading}
\subsection{Action Prediction}
\label{sec:action_prediction}
\vspace{\heading}

To learn from observational data, our algorithm must be able to estimate the actions used to transition between states. To do so, we train a model to estimate the action via supervised learning using the interaction data.
In particular, we train an inverse model, parameterized by $\invparam$, to calculate the action in the robot's action space, $\actint \in \mathcal{A}_{int}$, from a pair of invariant feature encodings, $(\featgen, \featgen')$.
Due to the isomorphism of the environment, we should expect to be able to predict actions for data from either MDP.
The inverse model is trained to minimize the mean squared error between the predicted action from the action-conditioned interaction data, $\predactint = \inv(\featint, \featint'; \invparam)$, and the corresponding true action $\actint$:
\vspace{-2mm}
\begin{equation}
    \mathcal{L}_a( \actint, \featint, \featint', \invparam) = \left\| \actint - \inv \left( \featint,\featint';\invparam\right) \right\|^2
    \label{eqn:inverse_model_loss}
\end{equation}
We then apply this trained inverse model to the action-free observation data.  We predict actions $\predactint = \inv(\featobs, \featobs';\invparam)$
and use them to train the reinforcement learning algorithm.
We predict actions in Line~\ref{alg:line:actions} and train the inverse model on Line~\ref{alg:line:inv} in Algorithm~\ref{alg:rl}.

\vspace{\heading}
\subsection{Reward Generation}
\vspace{\heading}
\label{sec:reward_generation}

One impediment to using observation data in reinforcement learning is that it lacks rewards.
We could use the same approach as in the previous section to predict rewards as well as actions, by training a reward model on top of the invariant features. However, this may perform poorly in sparse reward settings, where initial robot data contains no informative reward supervision.
In our implementation, we instead opt for a simple alternative to label the observation data with rewards using the methodology proposed by \citet{reddy2019sqil}.
The final timestep in a trajectory is assigned a large constant reward, $c_{large}$, while every other timestep is assigned a small constant reward, $c_{small}$:
\begin{equation}
R(\obsgen, \obsgen') = \begin{cases}
c_{large} &\obsgen' \text{ is terminal} \\
c_{small} &\obsgen'  \text{ is not terminal}
\end{cases}
\label{eqn:reward_generation}
\end{equation}
This encourages the agent to try to reach the states at the end of the observed trajectories, implicitly assuming that the observed trajectories are good.
We show that despite this assumption, the approach is still robust to observations from sub-optimal trajectories in Section~\ref{sec:suboptimal_demos}.
We expect this robustness arises because the observation data primarily acts to speed up exploration, while the robot gathers interaction data that can counteract any inaccuracies in the observation data.

\vspace{\heading}
\subsection{Domain Adaptation}
\label{sec:domain_adaptation}
\vspace{\heading}

As discussed in the previous sections, utilizing the observational data $(\obsobs, \obsobs')$ requires mapping it into an invariant representation $\featgen$. This is necessary both for training the inverse model and for utilizing the resulting tuples $(\featobs, \predactint, \featobs', \predrewardobs)$ alongside samples from the original MDP $(\featint, \actint, \featint', \rewardint)$. In this section, we describe how we can train such an invariant encoder, following the methodology in the domain adaptation literature~\citep{tzeng2017adversarial}.

We train our feature extractor, $\enc$,
to learn an encoded representation, $\featgen = \enc(\obsgen; \encparam)$, of an observation $\obsgen$. This encoded representation should contain all relevant information, while being invariant to the domain the observation originated from.
To train for such domain invariance, we separately train a discriminator, $\discrim$, 
to distinguish between encodings from the observational data, $\featobs = \enc(\obsobs;\encparam)$, and encodings from the interaction data, $\featint = \enc(\obsint; \encparam)$, and train the encodings in an adversarial manner, following work from \cite{tzeng2017adversarial}, according to the following loss:
\begin{align}
    \mathcal{L}_{\REV{dom}}(\obsint, \obsobs,\encparam, \discrimparam) = & \log \left( \discrim(\enc(\obsint;\encparam);\discrimparam) \right) +\log \left(1 - \discrim(\enc(\obsobs;\encparam);\discrimparam)\right)
    \label{eqn:adversarial_loss}
\end{align}
During training, the encoder attempts to minimize the discriminator's ability to correctly classify the domain of the encoded features, while the discriminator tries to maximize it.

We find that using the learned encoding, $\featgen$, for the inverse model while allowing the RL to use the raw observations offers the best performance.

\vspace{\heading}
\subsection{Joint Optimization}
\vspace{\heading}
We jointly optimize the domain adaptation loss with the inverse model loss, $\mathcal{L}_a$ and the optimization objective of the chosen reinforcement learning algorithm, $\mathcal{L}_\textsc{RL}$, according to the following objective:
\begin{align}
    \min_{\encparam, \invparam, \discrimparam}  
    \!\!\!\!\!\! \smash{\sum_{ \substack{ \{\obsobs, \obsobs'\} \in \poolobs,\\
    \{\obsint, \actint, \obsint'\} \in \poolint,\\ }}} \!\!\!\!\!\!
    c_2 \mathcal{L}_\textsc{RL} \nonumber
    &+  c_1\mathcal{L}_a( \actint, \enc(\obsint;\encparam), \enc(\obsint';\encparam), \invparam)
    \\ 
    &+ c_3 \max_{\discrimparam}  \mathcal{L}_{dom}(\obsint, \obsobs, \encparam, \discrimparam)
    \label{eqn:combined_loss}
\end{align}
where the term $\mathcal{L}_\textsc{RL}$ corresponds to the optimization objective of the chosen reinforcement learning algorithm. Constants $c_1$, $c_2$, and $c_3$ are hyperparameters that control the relative importance of each term.

All of the architecture details and hyperparameters are available in Appendix~\ref{app:hyperparams}.

\vspace{\heading}
\section{Experiments}
\vspace{\heading}
\label{sec:experiments}
The goal of our experiments is to evaluate whether RLV can enable more data-efficient reinforcement learning of robotic skills by incorporating video data without actions.
To this end, we study the following questions:
    \textbf{(1)} Can RLV learn from observations of sub-optimal policies?
    \textbf{(2)} Can RLV learn complex vision-based tasks with sparse rewards from observations and interaction?
    \textbf{(3)} Can RLV learn from videos with large amounts of domain shift, such as videos of humans?
We test these questions in three scenarios: learning simple control policies in the Acrobot-v1 environment~\cite{brockman2016openai},
learning simulated robotic behaviours using simulated observational data,
and learning simulated robotic behaviors using real-world human demonstrations.
In all experiments, we make use of soft-actor critic (SAC)~\cite{haarnoja2018soft} as the underlying reinforcement learning algorithm.  For all methods, we run five seeds and report the mean and standard error.\footnote{Videos and training code are available at our website: \url{https://sites.google.com/view/rl-with-videos}}

\subsection{Learning from Observations of sub-optimal policies}
\vspace{\heading}
\label{sec:suboptimal_demos}

We first evaluate whether RLV can learn from and improve upon observations from sub-optimal policies.
We perform these experiments on the Acrobot-v1 environment~\cite{brockman2016openai}.
To generate observations of sub-optimal trajectories, we train SAC~\cite{haarnoja2018sacapps} to convergence on the environment and take temporally consecutive blocks of observations from different points in training.
We discard the actions and the rewards in this data, so RLV only has access to the data it would have if it was learning from human observations.
We compare RLV to two prior methods that can leverage the observational data, ILPO~\cite{edwards2018imitating} and BCO~\cite{torabi2018behavioral}.
For both algorithms, we use the implementation provided by~\cite{edwards2018imitating} and the hyperparameters the authors tuned for the Acrobot-v1 environment.
We additionally compare to SAC~\cite{haarnoja2018soft,haarnoja2018sacapps}, a standard off-policy RL algorithm that is unable to make use of the action and reward free observational data.
This provides a direct comparison to RLV as we use the same implementation of SAC as the underlying RL algorithm in RLV.
For this environment, and the environments in Section~\ref{sec:exp2}, we disable the domain-adaptation portion of RLV by setting $c_3 = 0$.

\begin{figure}
\vspace{-0mm}
    \centering
    \begin{subfigure}{0.325\textwidth}
    \includegraphics[width=0.99\linewidth]{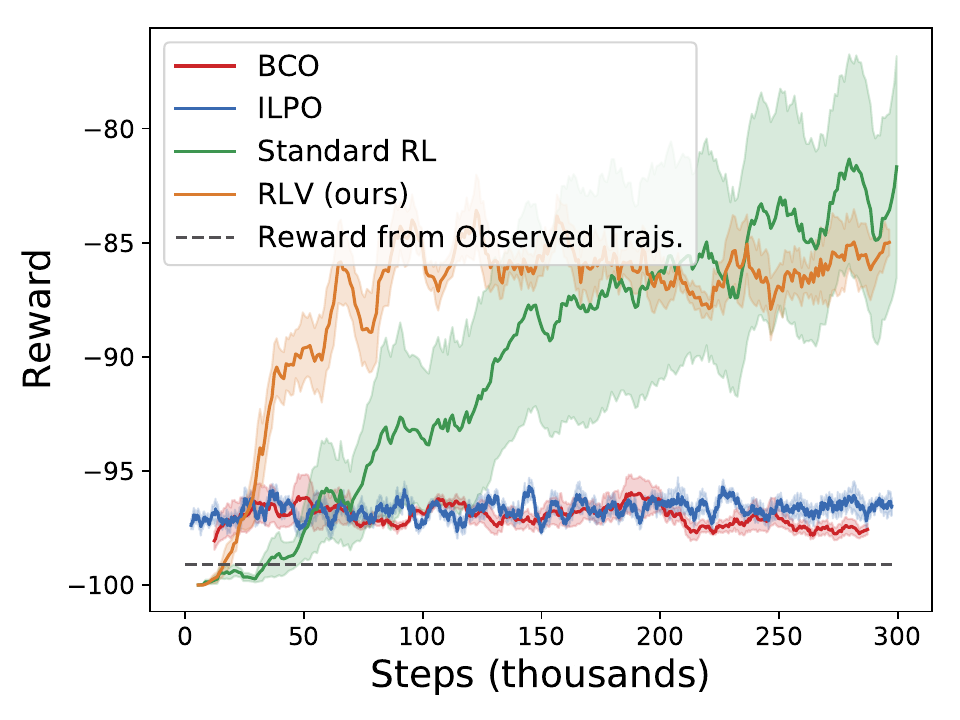}
\vspace{-6mm}
   \caption{\small  $\mathcal{D}_{obs}$ from \\SAC trained for 25K steps}
    \label{fig:acrobat_bad_demo}
    \end{subfigure}
    \begin{subfigure}{0.325\textwidth}
    \includegraphics[width=0.99\linewidth]{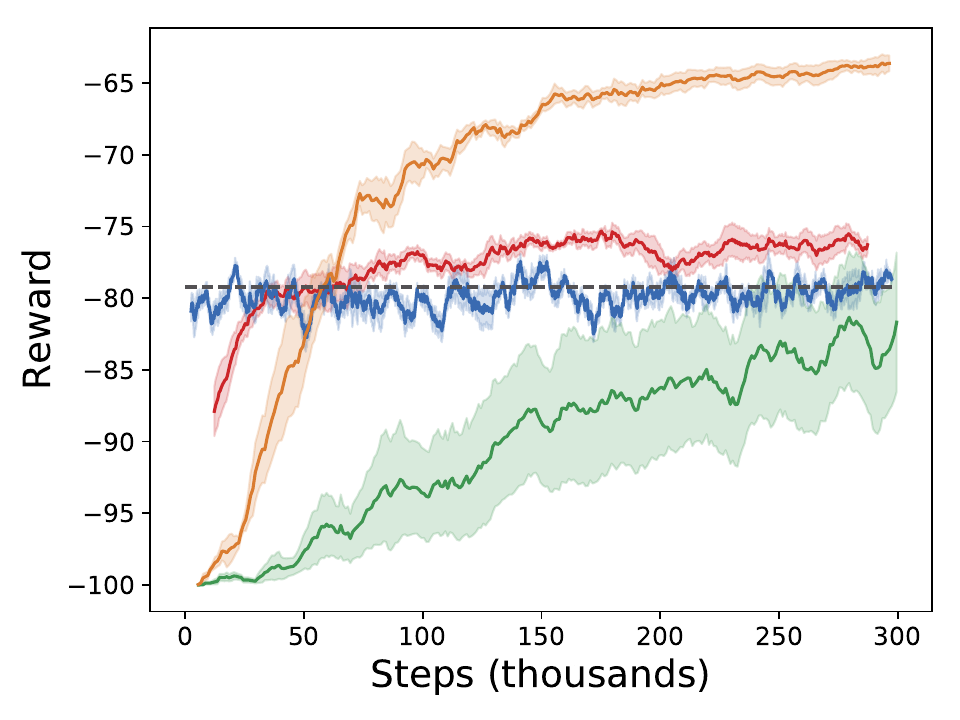}
\vspace{-6mm}
    \caption{\small $\mathcal{D}_{obs}$ from \\SAC trained for 225K steps}
    \label{fig:acrobot_med_demo}
    \end{subfigure}
    \begin{subfigure}{0.325\textwidth}
    \includegraphics[width=0.99\linewidth]{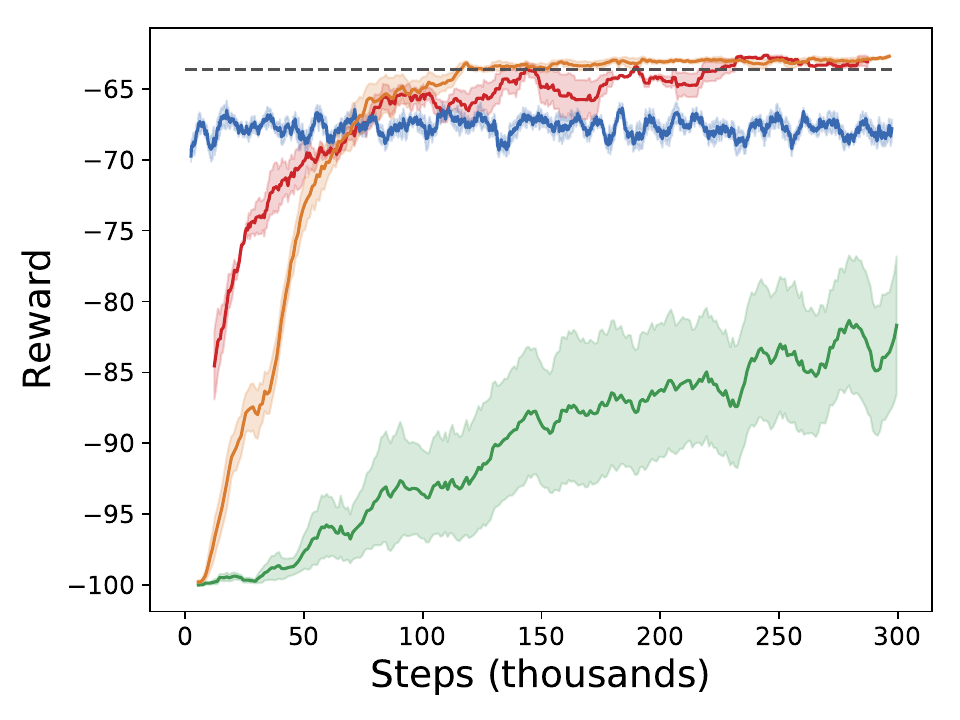}
\vspace{-6mm}
    \caption{\small $\mathcal{D}_{obs}$ from \\SAC trained for 975K steps}
    \label{fig:acrobot_good_demo}
    \end{subfigure}
    \caption{\small Performance on Acrobot with different qualities of observation data. RLV is generally able to achieve equal or higher final rewards than the competing methods while training with fewer samples.  The performance is especially notable with medium-quality observation data.}
    \label{fig:acrobot}
\end{figure}

The results for these experiments are shown in Figure~\ref{fig:acrobot}.  
We find that RLV consistently achieves higher rewards than ILPO~\cite{edwards2018imitating}, BCO~\cite{torabi2018behavioral}, and standard reinforcement learning~\cite{haarnoja2018soft}.  ILPO is more data-efficient but it is unable to improve upon observations from sub-optimal trajectories, while standard reinforcement learning trains requires more samples than RLV as it cannot leverage the offline observation data.
The performance improvements are most significant in Figure~\ref{fig:acrobot_med_demo}, where the observations came from a policy that achieved a medium level of performance.
Good performance with medium-quality data is particularly important since acquiring high-quality observation data is difficult and the domain gap will implicitly degrade their quality, while low quality observations contain little information so any method will learn slowly.

\vspace{\heading}
\subsection{Robotic Tasks}
\vspace{\heading}
\label{sec:exp2}



\begin{figure}
\newcommand{\pfwidth}{0.3\linewidth}
\centering
\hspace{0.04\textwidth}
    \begin{subfigure}{0.45\textwidth}
    
    \includegraphics[width=\pfwidth]{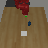}
    \includegraphics[width=\pfwidth]{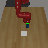}
    \includegraphics[width=\pfwidth]{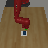}
\vspace{-2mm}
    \caption{\small Visual Pusher environment}
    \label{fig:push_forward_frames}
    \end{subfigure}
    \begin{subfigure}{0.45\textwidth}
    \includegraphics[width=\pfwidth]{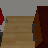}
    \includegraphics[width=\pfwidth]{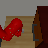}
    \includegraphics[width=\pfwidth]{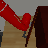}
 \vspace{-2mm}
    \caption{\small Visual Door Opening environment}
    \label{fig:door_pull_hook_frames}
    \end{subfigure}
    \vspace{-0.2cm}
    \caption{\small Randomly selected trajectories from the policy learned by RLV in different environments.  Both trajectories were successful.}
    \vspace{-5mm}
\end{figure}

\begin{figure}
    \centering
\vspace{-4mm}
    \begin{subfigure}{0.325\textwidth}
    \includegraphics[width=0.99\linewidth]{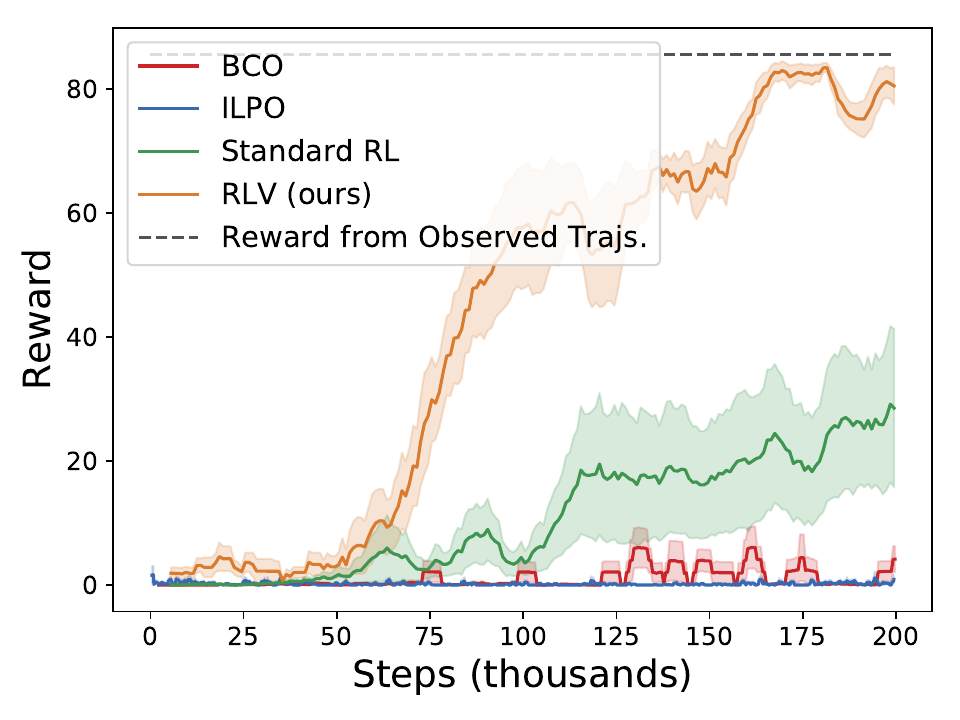}
\vspace{-6mm}
    \caption{\small State Pusher. }
    \label{fig:state_sawyer}
    \end{subfigure}
    \begin{subfigure}{0.325\textwidth}
    \includegraphics[width=0.99\linewidth]{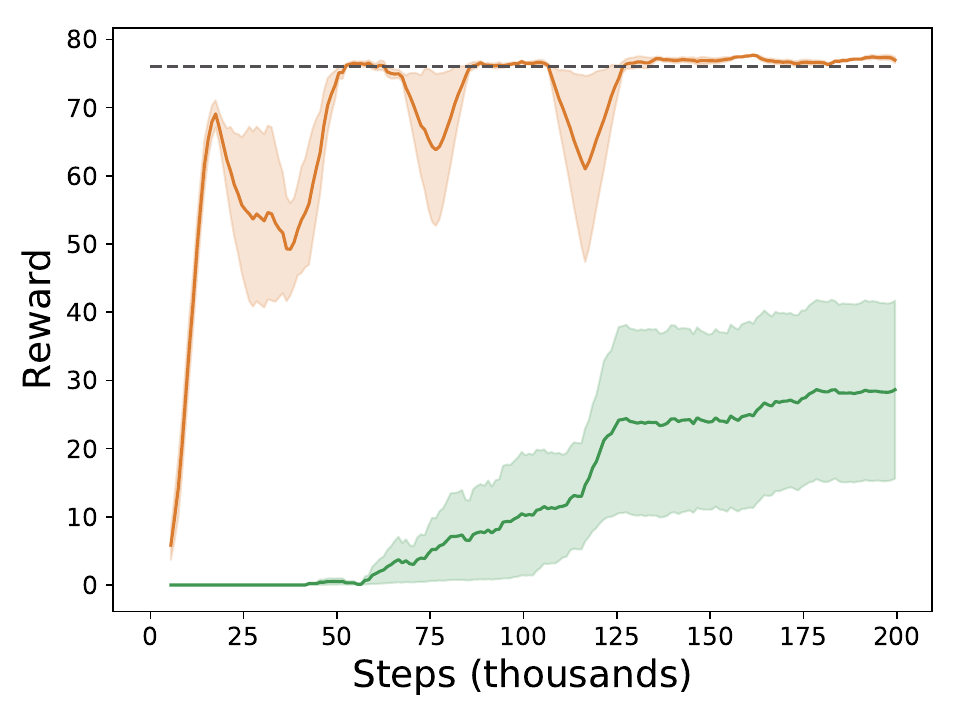}
\vspace{-6mm}
    \caption{\small Visual Door Opening. }
    \label{fig:door_pull_hook}
    \end{subfigure}
    \begin{subfigure}{0.325\textwidth}
    \includegraphics[width=0.99\linewidth]{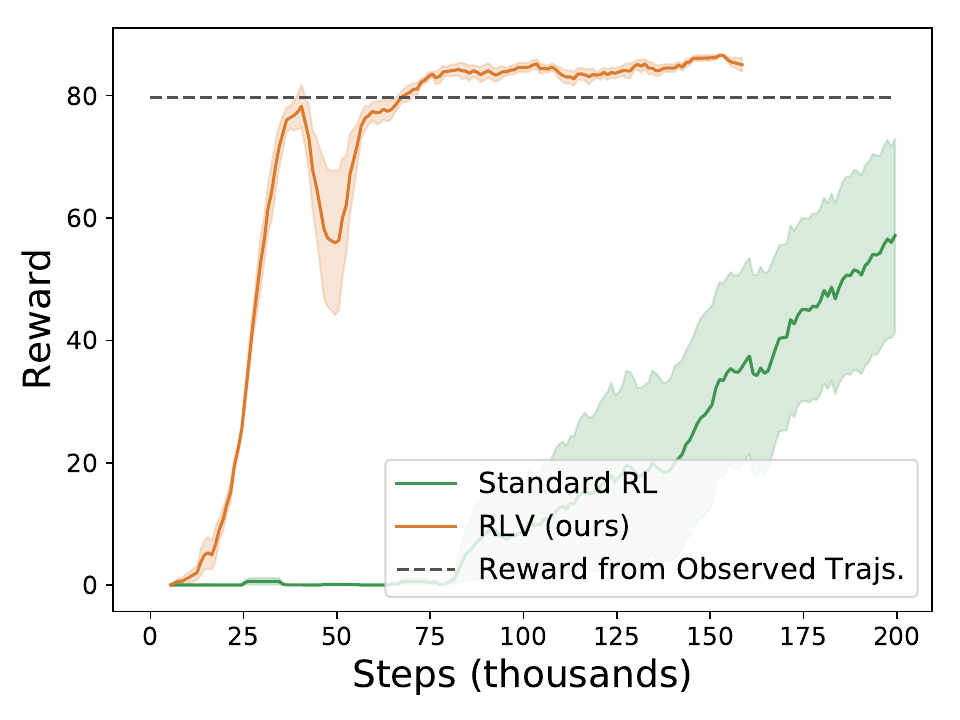}
\vspace{-6mm}
    \caption{\small Visual Pusher. }
    \label{fig:push_forward}
    \end{subfigure}
\vspace{-2mm}
    \caption{\small Rewards for the State Pusher, Visual Door Opening, and the Visual Pusher environments.  In both simulated environments, the agent trained with RLV requires fewer samples to solve the task than conventional reinforcement learning.}
    \label{fig:multiworld_image_environments}
\end{figure}

Next, we evaluate whether RLV can learn more complex robotic manipulation tasks using observations without domain shift.
We first examine a robotic pushing task~\cite{singh2019end}.  Unlike the Acrobot environment, which has a discrete action space with three possible values, the pushing task has an action space with two continuous dimensions, significantly increasing the difficulty of determining the actions that caused a sequence of observations.
The reward in this environment is sparse: it is one if the puck is within three centimeters of the goal location, and zero otherwise.
The observational data is from training SAC for 1M steps and taking observations of trajectories generated by the resulting policy.
We present results for the pushing task using state-based observations in Figure~\ref{fig:state_sawyer}.  RLV requires ~30\% as many samples to solve the task as
SAC, while BCO and ILPO fail to ever reliably solve the task,
as they struggle with the larger action space or more complicated dynamics.

We also investigate image based observations, using the Visual Pusher and the Visual Door Opening tasks from \cite{singh2019end}.
We again use a sparse binary reward function. The reward for the Visual Pusher is the same as the previous experiment, while the reward for the Visual Door Opening environment is zero if the door was not open to within a five degrees of the goal angle and one if the door is.
We were unable to train with BCO or ILPO to achieve non-zero rewards on tasks with image-based observations.
The results for these environments are shown in Figure~\ref{fig:multiworld_image_environments}.
We find that RLV requires 3x fewer samples as SAC in both environments. Further, on the Visual Door Opening environment in Figure~\ref{fig:door_pull_hook}, only three out of five of the training seeds of SAC converge to the optimal policy, while all of the training seeds of RLV reach the optimal policy.
Randomly sampled trajectories from the policies learned by RLV are shown in Figure~\ref{fig:push_forward_frames} for the Visual Pusher and in Figure~\ref{fig:door_pull_hook_frames} for the Visual Door Opening environment.

\vspace{\heading}
\subsection{Learning from Real-World Human Videos}
\label{sec:human_pushing}
\vspace{\heading}

\begin{figure}
\vspace{-4mm}
    \centering
    \hspace{0.02cm}
    \begin{subfigure}{0.38\textwidth}
        \centering
        \includegraphics[height=0.5\linewidth]{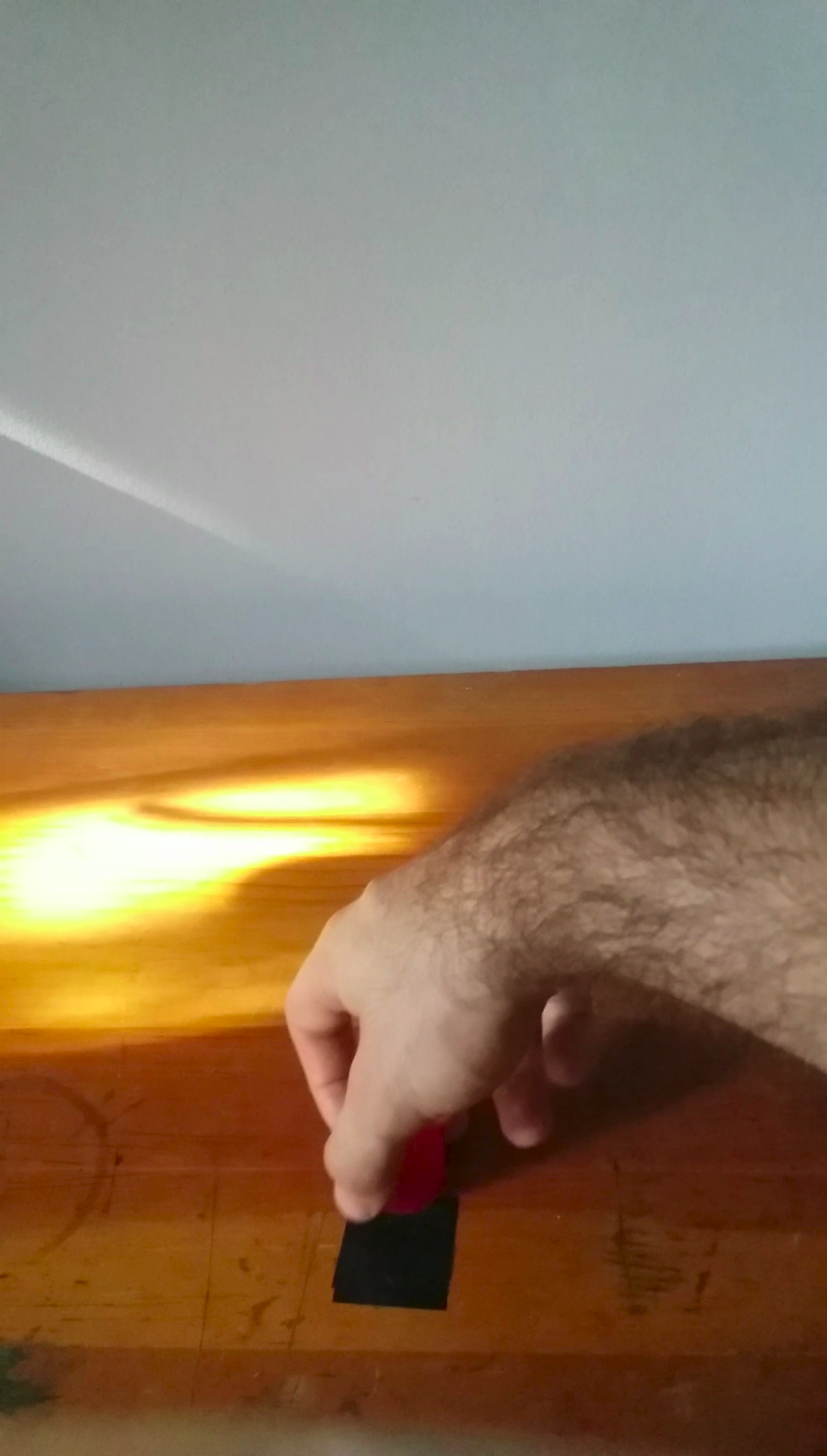}
        \includegraphics[height=0.5\linewidth]{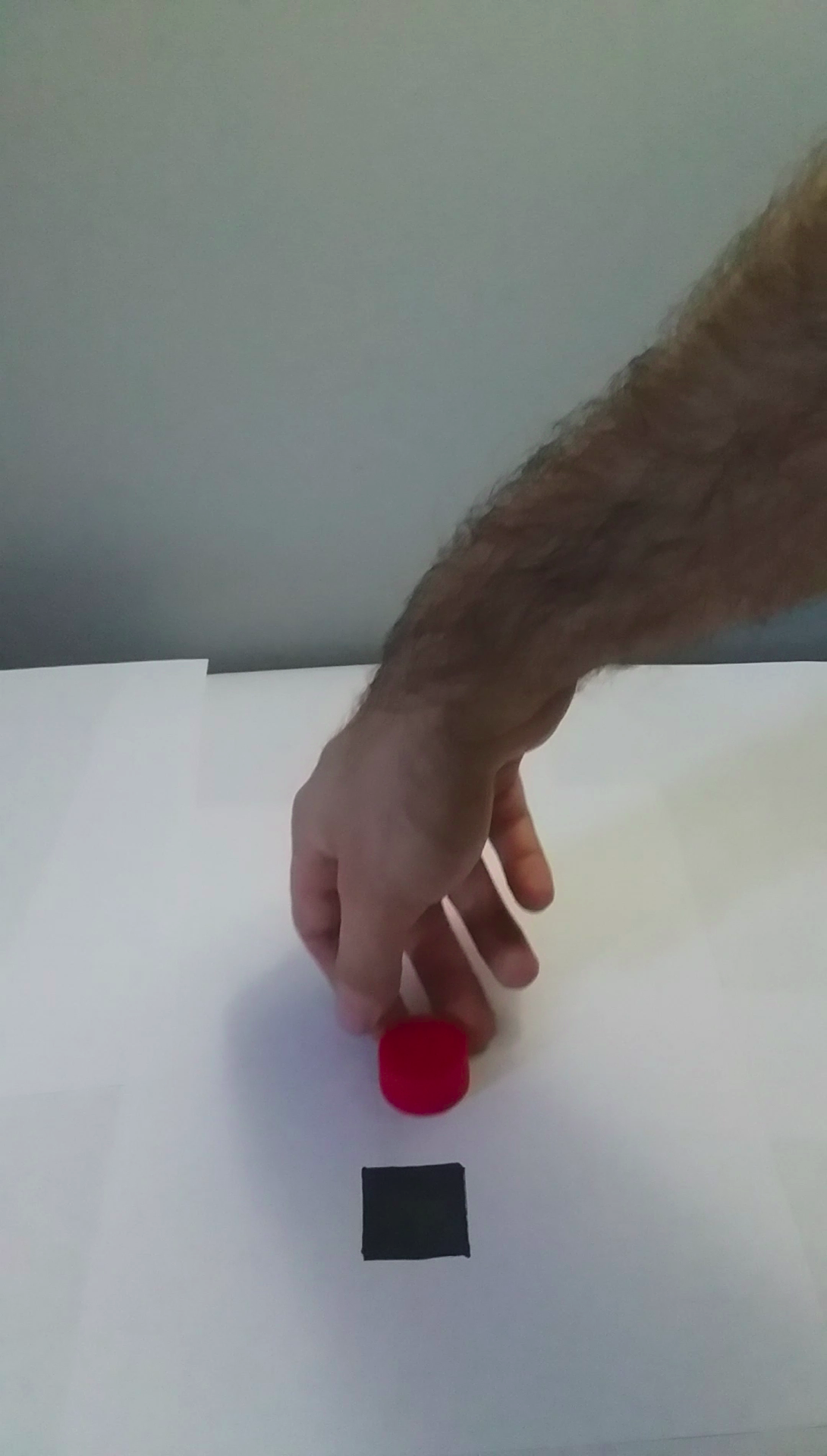}
        \includegraphics[height=0.5\linewidth]{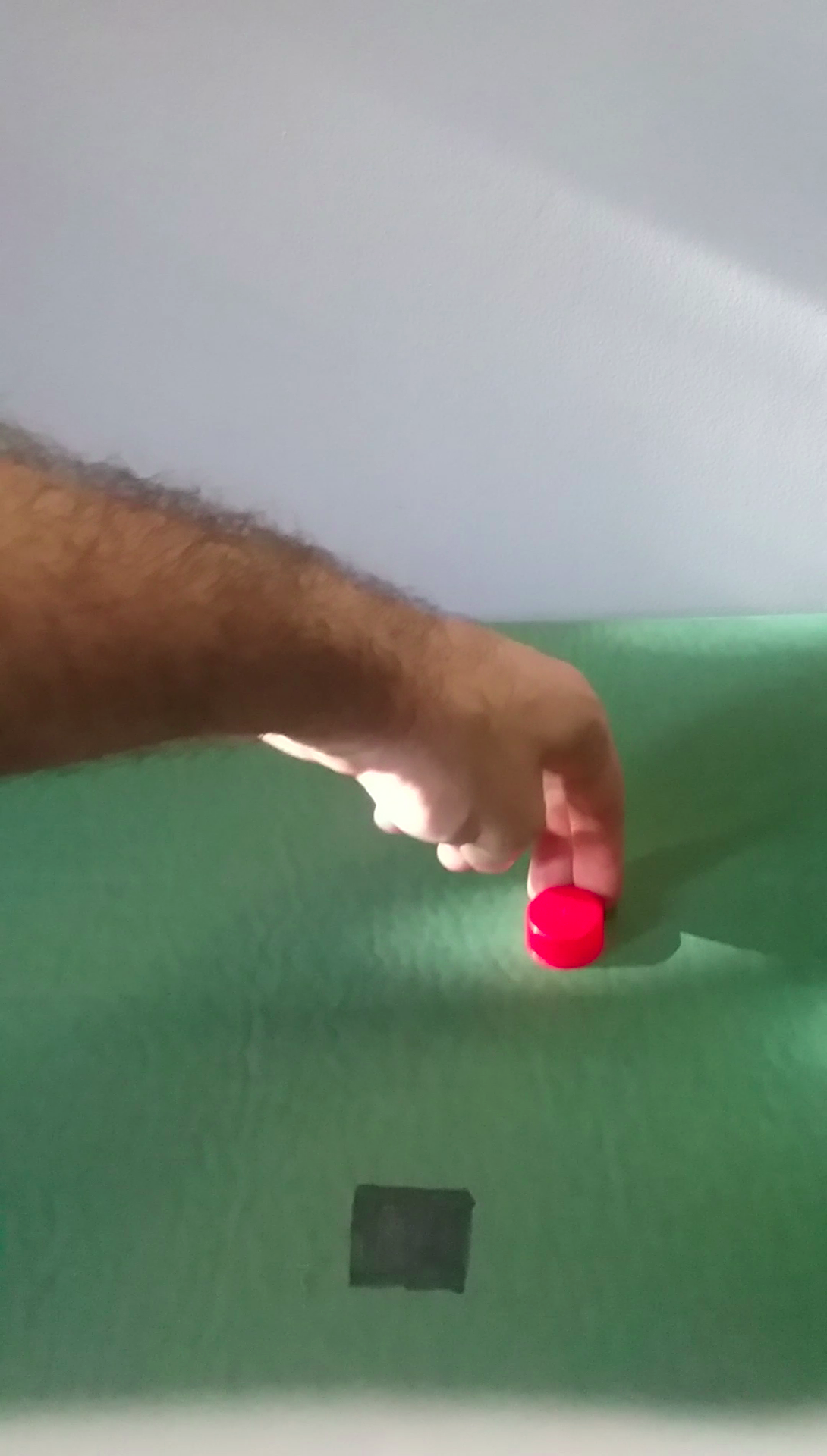}
        \caption{\small \centering Human observations}
        \label{fig:human_observation}
    \end{subfigure}
    \begin{subfigure}{0.22\textwidth}
        \centering
        \includegraphics[height=.88\linewidth]{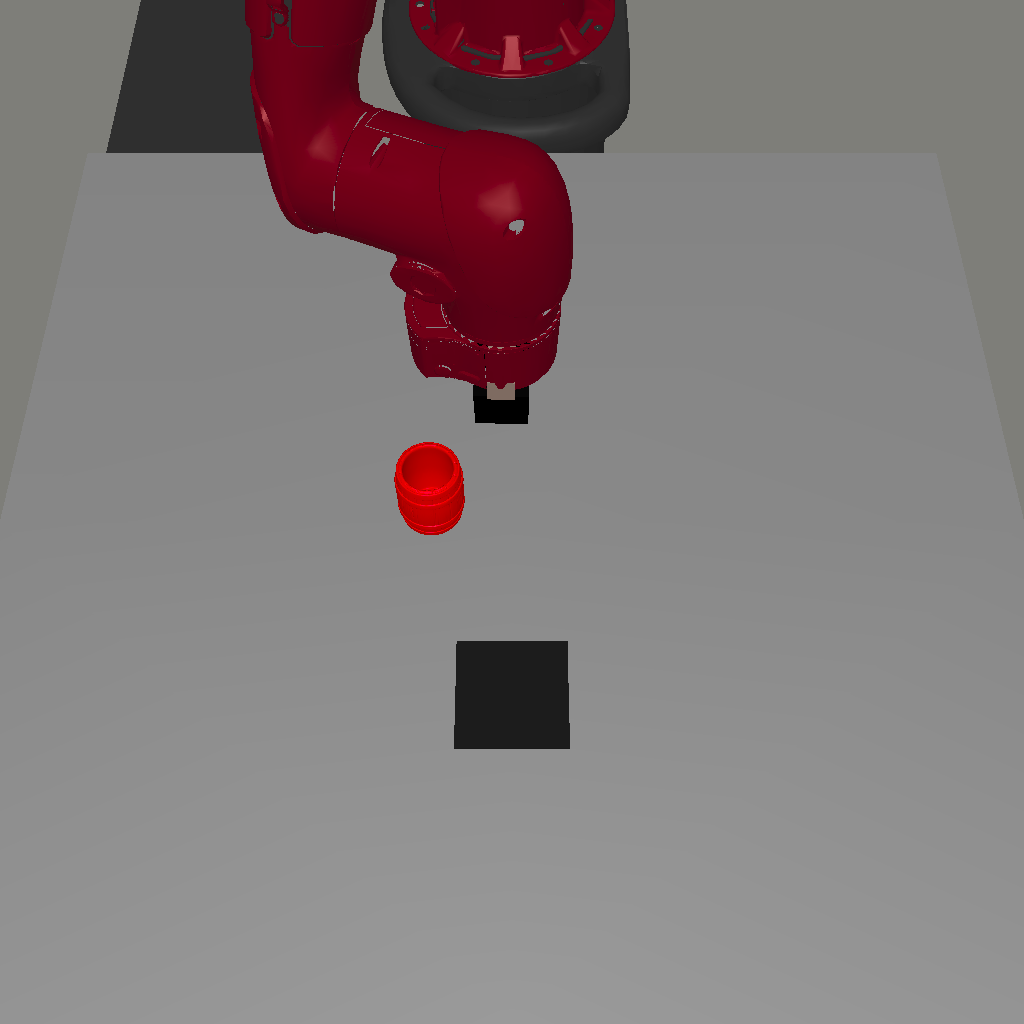}
        \caption{\small \centering Robot observations}
        \label{fig:humanlike_simulated}
    \end{subfigure}
    \begin{subfigure}{0.38\textwidth}
        \centering
        \includegraphics[height=0.6\linewidth]{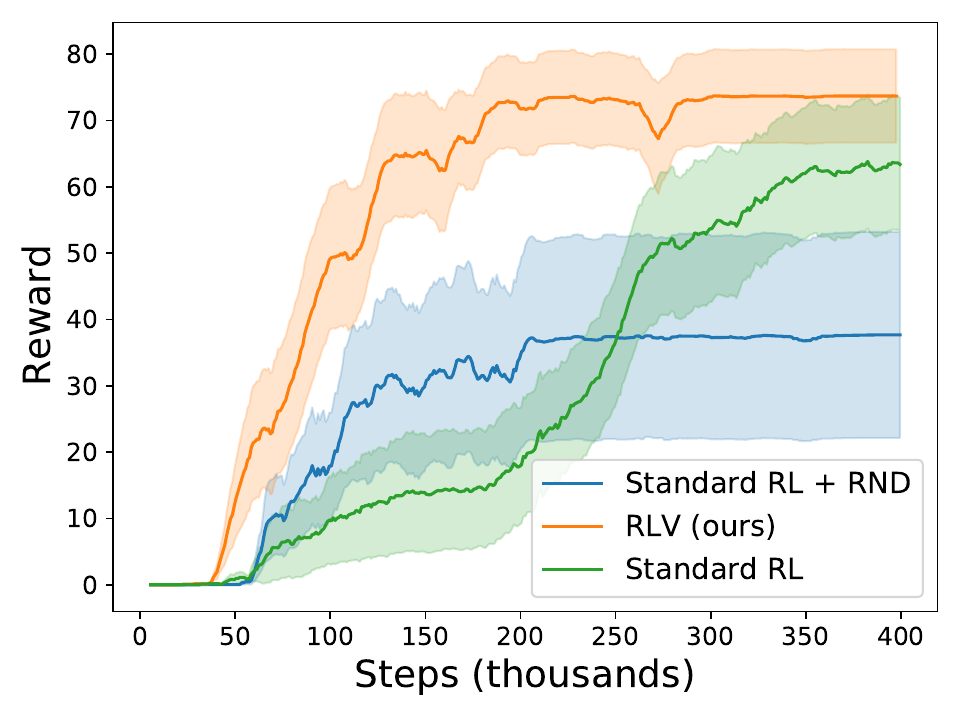}
\vspace{-2mm}
        \caption{\small \centering Training curves.}
        \label{fig:human_training_curves}
    \end{subfigure}
\vspace{-2mm}
    \caption{\small Training curves and example images for learning pushing with human observations.  Our model is able to leverage videos of humans to solve the task with significantly fewer samples than conventional reinforcement learning.}
\vspace{-5mm}
\end{figure}

\begin{figure}[t]
\vspace{-6mm}
    \centering
    \hspace{0.02cm}
    \begin{subfigure}{0.38\textwidth}
        \centering
        \includegraphics[height=0.5\linewidth]{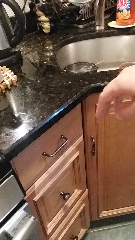}
        \includegraphics[height=0.5\linewidth]{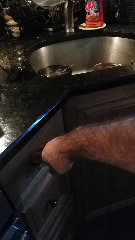}
        \includegraphics[height=0.5\linewidth]{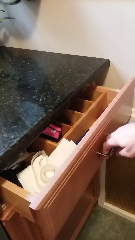}
        \caption{\small \centering Human observations}
        \label{fig:human_drawer_observation}
    \end{subfigure}
    \begin{subfigure}{0.22\textwidth}
    \centering
        \includegraphics[height=0.88\linewidth]{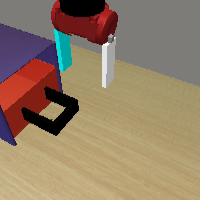}
        \caption{\small \centering Robot observations}
        \label{fig:drawer_simulated}
    \end{subfigure}
    \begin{subfigure}{0.38\textwidth}
    \centering
        \includegraphics[height=0.6\linewidth]{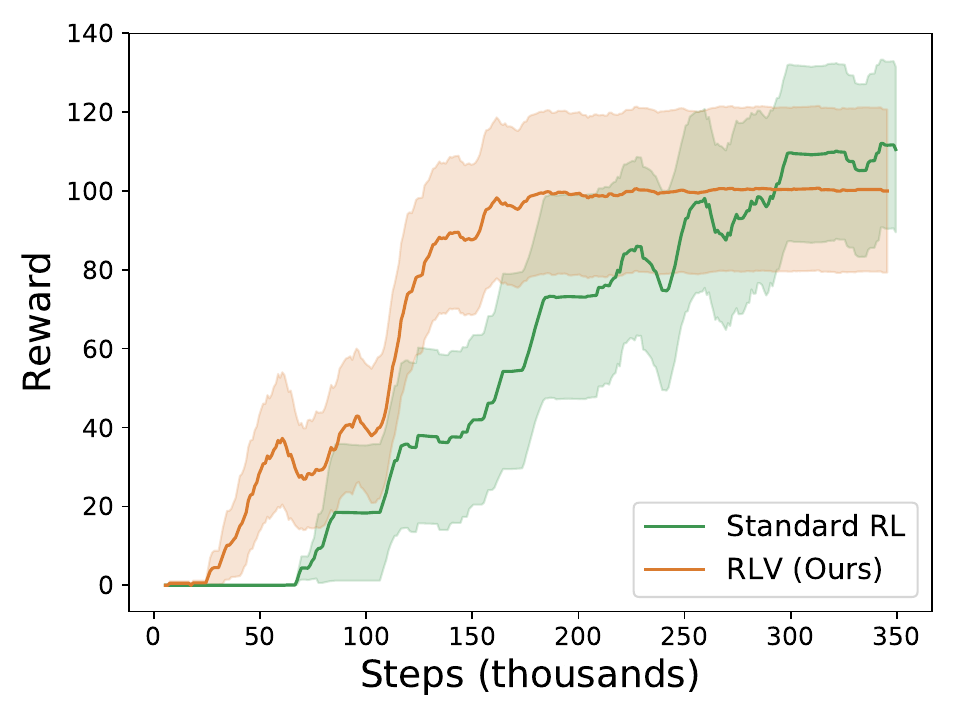}
\vspace{-2mm}
        \caption{\small \centering Training curves.}
         \vspace{-1mm}
        \label{fig:human_training_curves_drawer}
    \end{subfigure}
\vspace{-1mm}
    \caption{\small Training curves and example images for learning drawer opening with human observations.  Our model is able to leverage videos of humans to solve the task with significantly fewer samples than conventional reinforcement learning.}
    \vspace{-6mm}
\end{figure}

We next seek to study if RLV can learn robotic policies from videos of humans.
Given videos of humans completing the task in the real world,
we train an agent to complete a similar task in simulation.
In addition to learning from observations without actions or rewards, the agent must also overcome the domain shift between real and simulated images and between humans and robots. 
We consider two tasks, pushing an object and opening a drawer.
Example observation from the human domain are shown in Figure~\ref{fig:human_observation} for the pushing task and Figure~\ref{fig:human_drawer_observation} for the drawer opening task.  Example observations from the robot domain are shown in Figure~\ref{fig:humanlike_simulated} for the pushing task and Figure~\ref{fig:drawer_simulated} for the drawer opening task.
Note that there is considerable domain shift in color, lighting conditions, object positions, and embodiment.  More information on the datasets are available in Appendix~\ref{sec:dataset}.
For the pushing task, our simulated environment is a recolored \REV{and more difficult} version of the previous Visual Pusher environment \cite{singh2019end} using the same sparse reward function as our previous experiments. 
For the drawer opening task, our simulated environment is a modified version of the drawer opening task from Meta-World~\cite{yu2019meta} with a reduced action space and a more difficult initial configuration.

The results for different approaches are shown in Figure~\ref{fig:human_training_curves} for the pushing task and Figure~\ref{fig:human_training_curves_drawer} for the drawer opening task.  RLV is able to leverage the human observations to significantly speed up training, requiring only 50\% of the samples that SAC requires to match its performance in both tasks. \REV{Further, to evaluate whether a similar improvement may be achieved with a task-agnostic exploration technique, we compare to a version of SAC that uses the exploration bonus from RND \cite{burda2018exploration}. We observe that RND learns faster than default RL, but still achieves inferior learning speed and final performance compared to RLV.}

\vspace{\heading}
\subsection{Ablations over action and reward prediction}
\vspace{\heading}

\begin{wrapfigure}{R}{0.34\textwidth}
\vspace{-12mm}
    \centering
    \begin{subfigure}{0.99\linewidth}
    \includegraphics[width=0.99\linewidth]{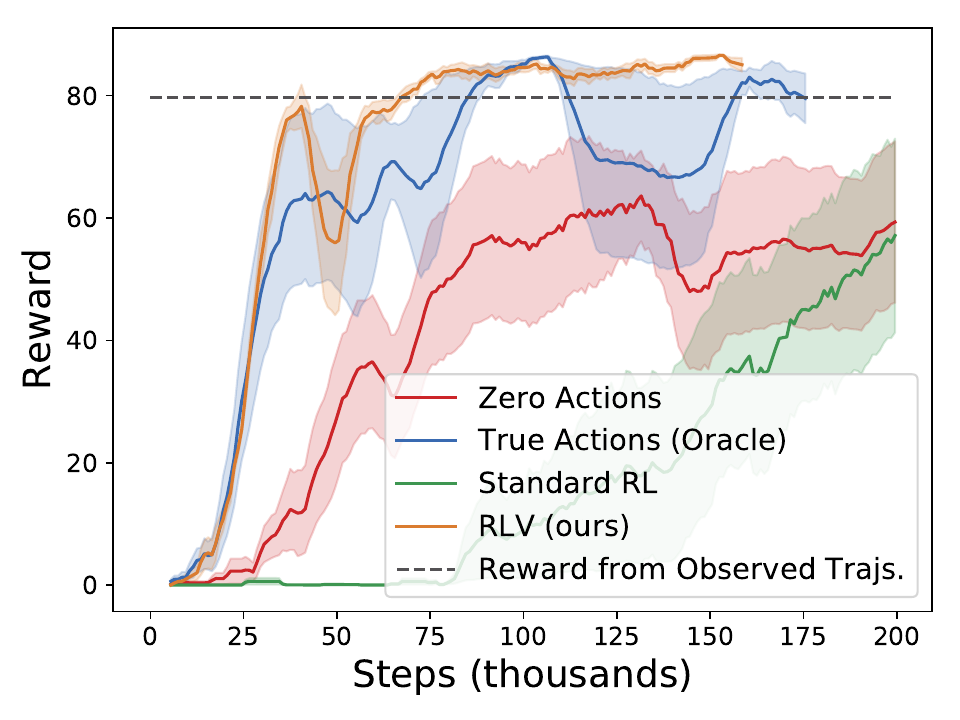}
\vspace{-6mm}
    \caption{\centering RLV with different action predictors}
    \label{fig:action_ablation}
    \end{subfigure}
    
    \begin{subfigure}{0.99\linewidth}
    \includegraphics[width=0.99\linewidth]{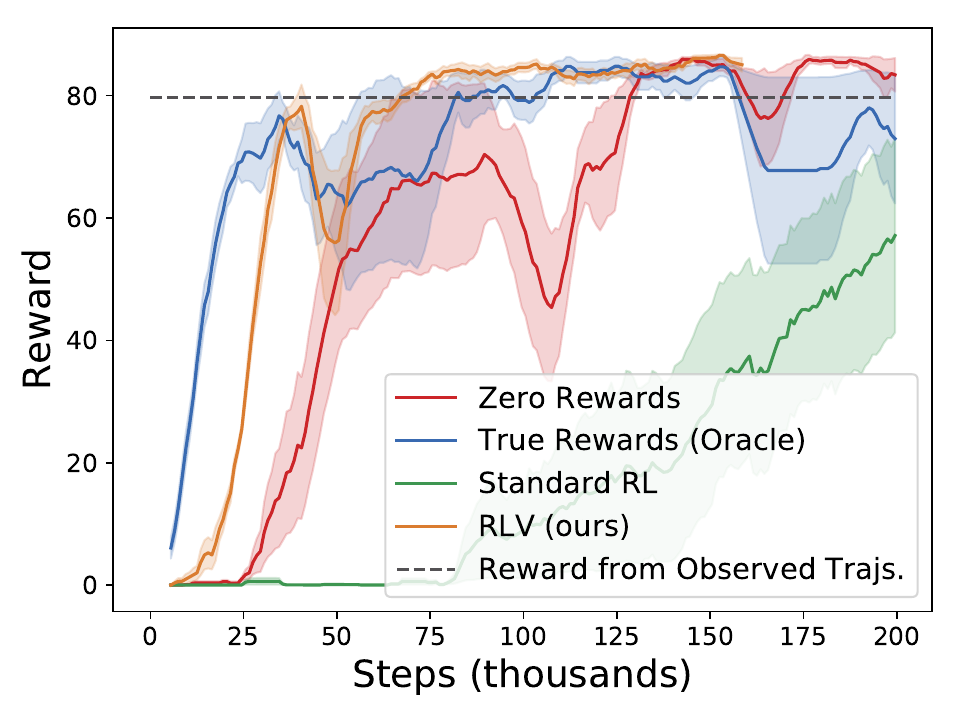}
\vspace{-6mm}
    \caption{\centering RLV with different reward predictors}
    \label{fig:reward_ablation}
    \end{subfigure}
\vspace{-2mm}
    \caption{\small Ablations in the visual pusher environment over different action and reward predictors for the observation data.  Even when the predictor always predicts zeros for the actions or rewards, our approach is more data-efficient than standard RL, while our method of estimating the actions and rewards performs comparably to using ground truth values.}
    \label{fig:ablations}
\vspace{-7mm}
\end{wrapfigure}
We perform ablations over the actions and the rewards used for the observation data in the Visual Pusher environment.
For rewards and actions, we compare using the ground truth values, using the values estimated by our approach, and using all zeros.
The results are shown in Figure~\ref{fig:ablations}.
For the actions, both our approach and using the ground truth actions perform well, while using zero actions initially performs better than the baseline reinforcement learning, but converges to a sub-optimal final score.
This indicates that even poor estimates of the actions can speed up training. However, better estimates are required to consistently find the optimal policy and the actions estimated by the inverse model are of sufficient quality to accomplish this.
For the rewards, both our approach and using the ground truth rewards perform comparably well, while using zero rewards underperforms but still converges to the optimal solution.
This result likely relies on the fact that the ground truth reward is mostly zeros except for at the end, making it such the zero reward contains little incorrect information.
This result generally suggests that RLV is fairly robust to the choice of reward labels.

\vspace{\heading}
\section{Conclusion}
\vspace{\heading}

We present reinforcement learning with videos, a framework for learning robotic policies using both online robotic interaction and offline observations of humans, incorporating the latter data by addressing the lack of actions and rewards and the domain shift with simple mechanisms.
By leveraging observations of humans, this framework is able to learn significantly more efficiently than standard reinforcement learning methods that cannot readily leverage that data. It also outperforms prior approaches for imitation learning from observation, as it is able to improve over the demonstrator performance, tackle complex vision-based tasks, and handle large domain shift.
Future work includes generalizing to even more visually complex scenes and handling more natural and diverse human demonstrations, while increasing the complexity of the learned behaviors.
Overcoming these challenges will allow reinforcement learning to leverage the vast quantities of diverse and interesting observations of humans, making robotic agents both easier to train and more capable of solving challenging tasks.


	



	


\clearpage
\acknowledgments{
We would like to thank Kenneth Chaney, Wendelyn Bolles, Minttu Alakuijala, and our anonymous internal and external reviewers.
This work was supported by ARL RCTA W911NF-10-2-0016, ARL
DCIST CRA W911NF-17-2-0181, ONR grant N00014-20-1-2675, and by Honda Research Institute.}


{\small
\bibliography{main}  
}

\newpage
\appendix

\section{Revisions from CoRL Version}
\label{sec:changes}
This paper was revised from the initially published version to correct for some inaccuracies.  Due to a bug in the code, the initial paper incorrectly claimed that RLV utilized paired data.  In reality, RLV does not require any paired data between the human and robot domains and all experiments in the original paper were performed without making use of the paired data.

This paper has been corrected by removing references to the paired data, making it so the text of the paper accurately describes the implemented algorithm.

Additionally, rather than sharing one feature extractor between all networks, the policy network, the Q functions, and the action encoder each make use of their own feature extractor.

\section{Data and Qualitative Results on Supplementary Website}

Dataset downloads and videos of our qualitative results are hosted on our website:
\url{https://sites.google.com/view/rl-with-videos}.

\section{COVID-19 Statement}
Due to COVID-19, we were unable to evaluate RLV on a real robot. However, the experiments in the paper provide significant evidence that the algorithm would be successful on a real robot.  First, while there will be domain shift between the human observations and the real robot, this domain shift will likely be no larger than the domain shift betwee real human videos and simulated robot observations in the experiments in Section 4.3. Second, the simulated results show good performance with raw visual observations and without the need for a shaped reward function, both of which make it practical to deploy to a real robot without significant instrumentation of the environment. Third, our results suggest that the simulated robot learns within 75,000 steps or 750 episodes, an amount of data that is practical to collect on a real robot. Finally, RLV builds upon the SAC algorithm, which has been previously demonstrated on a real robot with raw pixel observations~\cite{haarnoja2018sacapps,singh2019end}.

\section{Architecture Details and Hyperparameters}
\label{app:hyperparams}

We instantiate our model with deep neural networks. 
We use SAC as the underlying RL algorithm in RLV.
Hyperparameters for our different experiments are shown in Tables~\ref{tab:general_hyperparameters}-\ref{tab:human_hyperparameters}.

We used the default hyperparameters for SAC wherever possible, and performed a hyperparameter sweep to find the remaining parameters.
The most sensitive hyperparameters are the ones in Table~\ref{tab:human_hyperparameters}, which control the balance between domain invariance and the other losses for training the feature representation.
The reward generation parameter, $c_{large}$, is robust, capable of using the same value in different environments with different reward scales.  The other reward generation parameter, $c_{small}$, is less robust, but can simply be set to the reward for taking a random step at the starting state in each environment.
The network architecture hyperparameters are shown in Tables~\ref{tab:state_hyperparameters} and~\ref{tab:image_hyperparameters}; these hyperparameters have a range of viable values.

\begin{table}[h]
    \centering
    \begin{tabular}{|c|c|}
    \hline
        Name & Value \\
    \hline
        Learning Rate & 3e-4 \\
        Num Initial Exploration Steps & 1000 \\
        Batch Size & 256 \\
        Optimizer & ADAM~\cite{kingma2014adam} \\
        Nonlinearity & ReLU \\
        Gradient Steps per environment step & 1 \\
    \hline
    \end{tabular}
    \caption{General Hyperparameters.  These hyperparameters are used for RLV and for SAC in all environments.}
    \label{tab:general_hyperparameters}
\end{table}

\begin{table}[h]
    \centering
    \begin{tabular}{|c|c|c|}
    \hline
        Hyperparameter & Acrobot Environment & MuJoCo Environments \\
    \hline
        $c_{large}$ & 10.0 & 10.0\\
        $c_{small}$ & -1.0 & 0.0\\
    \hline
    \end{tabular}
    \caption{Reward generation parameters.  These hyperparameters were used to generate the rewards for RLV in different environments.  The MuJoCo environments include the State Pusher, Visual Pusher, Visual Door Opening, and the visual drawer opening environments.}
    \label{tab:reward_hyperparameters}
\end{table}

\begin{table}[h]
    \centering
    \begin{tabular}{|c|c|}
    \hline
        Name & Value \\
    \hline
        Q Function Fully Connected Layer Features & [256, 256]\\
        Policy Network Fully Connected Layer Features & [256, 256]\\
        Inverse Model Fully Connected Layer Features & [64, 64, 64] \\
    \hline
    \end{tabular}
    \caption{State Observation Architecture Hyperparameters.  These hyperparameters are used when the agent receives state based observations (Acrobot-v0 and State Pusher).}
    \label{tab:state_hyperparameters}
\end{table}

\begin{table}[h]
    \centering
    \begin{tabular}{|c|c|}
    \hline
        Name & Value \\
    \hline
        Feature Extractor Conv Number of Filters & [16, 16, 32]\\
        Feature Extractor Conv Filter Size & 5\\
        Input image shape & [48, 48, 3]\\
        Pooling Type & MaxPool \\
        Pooling Stride & 2 \\
        Q Function Fully Connected Layer Features & [512, 256, 256]\\
        Policy Network Fully Connected Layer Features & [512, 256, 256] \\
        Inverse Model Fully Connected Layer Features & [64, 64, 64] \\
    \hline
    \end{tabular}
    \caption{Image Observation Architecture Hyperparameters.  These hyperparameters are used when the agent receives image based observations (visual pusher, visual door opening, visual drawer opening).}
    \label{tab:image_hyperparameters}
\end{table}

\begin{table}[b]
    \centering
    \begin{tabular}{|c|c|}
    \hline
        Name & Value \\
    \hline
        $c_1$ & 1 \\
        $c_2$ & 1 \\
        $c_3$ & 0.001 \\
        Discriminator Fully Connected Layer Features & [64, 64, 64] \\
        Discriminator learning rate & 3e-8\\
        \REV{Paired data weight} & 1e-6\\
    \hline
    \end{tabular}
    \caption{Human Observation Hyperparameters.  These hyperparameters are used for RLV learning from human demonstrations.}
    \label{tab:human_hyperparameters}
\end{table}

For learning from observational data of humans, we augmented the images with random crops for both SAC and RLV following \cite{laskin2020reinforcement,kostrikov2020image}.
We pad the image on all sides with four black pixels, then randomly crop an image of the original size from the padded image.

While ILPO and BCO are designed for tasks with discrete actions, we were able to make them run on the robotic pushing task by discretizing the action space. 
We ran a hyperparamter sweep to determine the correct resolution of the discretization, selecting a discretization that broke the two-dimensional action space of the State Pusher environment into 49 discrete bins.

\section{Human Dataset}
\label{sec:dataset}
For our experiments in Section~\ref{sec:human_pushing}, we collected a dataset of a human pushing a puck \REV{and of a human opening a drawer}.
The \REV{pushing component of the} dataset is composed of 198 videos, totaling 806 seconds.  The data includes four backgrounds, two hands, two pucks, and a variety of lighting conditions.
\REV{The drawer opening component of the dataset is composed of 108 videos, totaling 376 seconds.  The data includes two drawers, one hand, and a variety of lighting conditions.}





\begin{figure}
\newcommand{\humanwidth}{0.11\linewidth}
    \centering
    \includegraphics[width=\humanwidth]{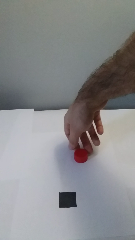}
    \includegraphics[width=\humanwidth]{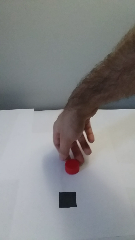}
    \includegraphics[width=\humanwidth]{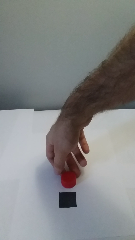}
    \includegraphics[width=\humanwidth]{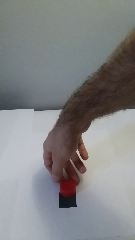}
    \includegraphics[width=\humanwidth]{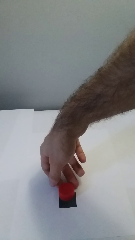}
    \includegraphics[width=\humanwidth]{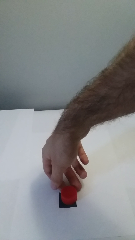}
    \includegraphics[width=\humanwidth]{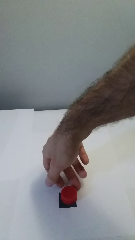}
    \includegraphics[width=\humanwidth]{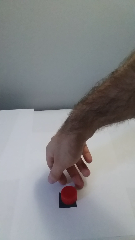}

    \includegraphics[width=\humanwidth]{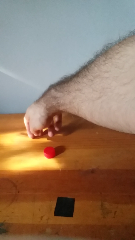}
    \includegraphics[width=\humanwidth]{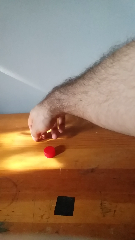}
    \includegraphics[width=\humanwidth]{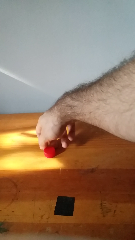}
    \includegraphics[width=\humanwidth]{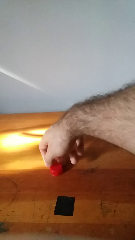}
    \includegraphics[width=\humanwidth]{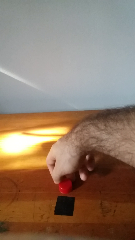}
    \includegraphics[width=\humanwidth]{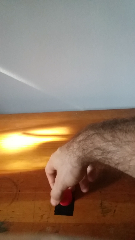}
    \includegraphics[width=\humanwidth]{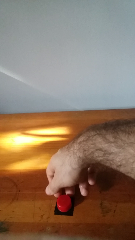}
    \includegraphics[width=\humanwidth]{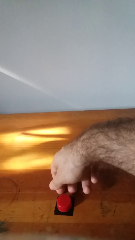}
    
    \includegraphics[width=\humanwidth]{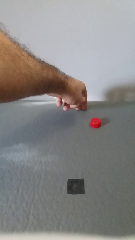}
    \includegraphics[width=\humanwidth]{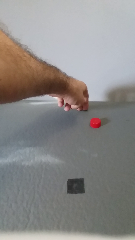}
    \includegraphics[width=\humanwidth]{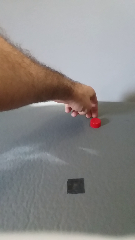}
    \includegraphics[width=\humanwidth]{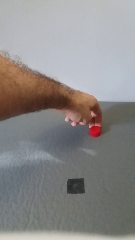}
    \includegraphics[width=\humanwidth]{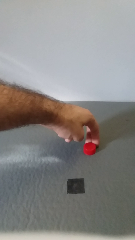}
    \includegraphics[width=\humanwidth]{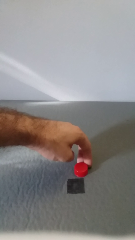}
    \includegraphics[width=\humanwidth]{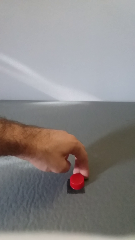}
    \includegraphics[width=\humanwidth]{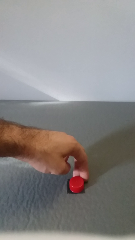}
    
    \includegraphics[width=\humanwidth]{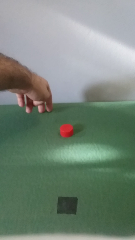}
    \includegraphics[width=\humanwidth]{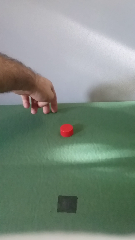}
    \includegraphics[width=\humanwidth]{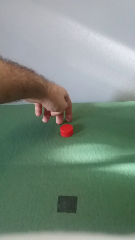}
    \includegraphics[width=\humanwidth]{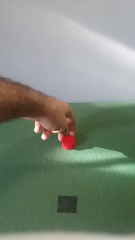}
    \includegraphics[width=\humanwidth]{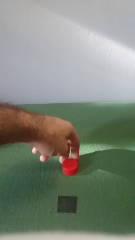}
    \includegraphics[width=\humanwidth]{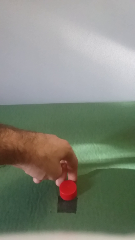}
    \includegraphics[width=\humanwidth]{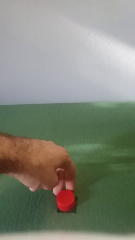}
    \includegraphics[width=\humanwidth]{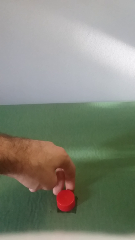}

    \caption{Example images from the human pushing observation data used by RLV.}
    \label{fig:human_dataset_appendix}
\end{figure}
\begin{figure}
\newcommand{\postwidth}{0.11\linewidth}
    \centering
    \scalebox{1}[-1]{\includegraphics[width=\postwidth]{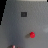}}
    \scalebox{1}[-1]{\includegraphics[width=\postwidth]{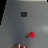}}
    \scalebox{1}[-1]{\includegraphics[width=\postwidth]{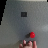}}
    \scalebox{1}[-1]{\includegraphics[width=\postwidth]{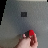}}
    \scalebox{1}[-1]{\includegraphics[width=\postwidth]{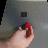}}
    \scalebox{1}[-1]{\includegraphics[width=\postwidth]{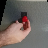}}
    \scalebox{1}[-1]{\includegraphics[width=\postwidth]{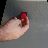}}
    \scalebox{1}[-1]{\includegraphics[width=\postwidth]{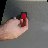}}
    
    \scalebox{1}[-1]{\includegraphics[width=\postwidth]{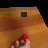}}
    \scalebox{1}[-1]{\includegraphics[width=\postwidth]{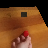}}
    \scalebox{1}[-1]{\includegraphics[width=\postwidth]{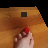}}
    \scalebox{1}[-1]{\includegraphics[width=\postwidth]{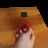}}
    \scalebox{1}[-1]{\includegraphics[width=\postwidth]{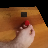}}
    \scalebox{1}[-1]{\includegraphics[width=\postwidth]{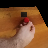}}
    \scalebox{1}[-1]{\includegraphics[width=\postwidth]{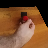}}
    \scalebox{1}[-1]{\includegraphics[width=\postwidth]{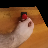}}
    
    \scalebox{1}[-1]{\includegraphics[width=\postwidth]{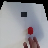}}
    \scalebox{1}[-1]{\includegraphics[width=\postwidth]{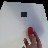}}
    \scalebox{1}[-1]{\includegraphics[width=\postwidth]{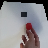}}
    \scalebox{1}[-1]{\includegraphics[width=\postwidth]{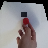}}
    \scalebox{1}[-1]{\includegraphics[width=\postwidth]{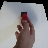}}
    \scalebox{1}[-1]{\includegraphics[width=\postwidth]{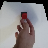}}
    \scalebox{1}[-1]{\includegraphics[width=\postwidth]{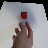}}
    \scalebox{1}[-1]{\includegraphics[width=\postwidth]{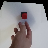}}
    
    \caption{Example post-processed images from the human observation data used by RLV.}
    \label{fig:human_dataset_post_appendix}
\end{figure}
\begin{figure}
\newcommand{\humanwidth}{0.11\linewidth}
    \centering
    \includegraphics[width=\humanwidth]{figures/drawer_human_frames/im_000_0000.png}
    \includegraphics[width=\humanwidth]{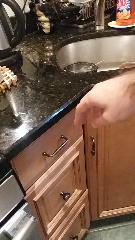}
    \includegraphics[width=\humanwidth]{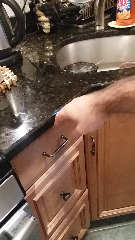}
    \includegraphics[width=\humanwidth]{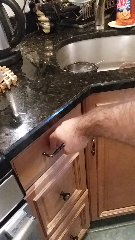}
    \includegraphics[width=\humanwidth]{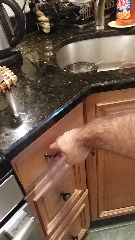}
    \includegraphics[width=\humanwidth]{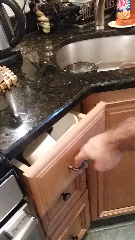}
    \includegraphics[width=\humanwidth]{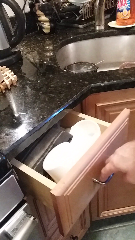}
    \includegraphics[width=\humanwidth]{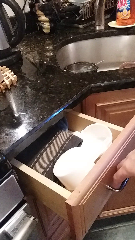}
    
    \includegraphics[width=\humanwidth]{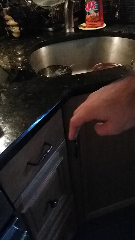}
    \includegraphics[width=\humanwidth]{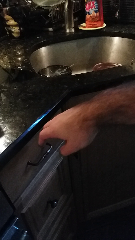}
    \includegraphics[width=\humanwidth]{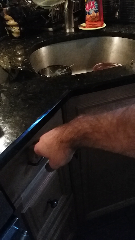}
    \includegraphics[width=\humanwidth]{figures/drawer_human_frames/im_001_0039.png}
    \includegraphics[width=\humanwidth]{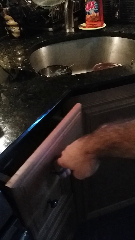}
    \includegraphics[width=\humanwidth]{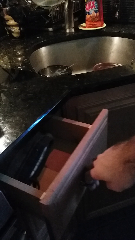}
    \includegraphics[width=\humanwidth]{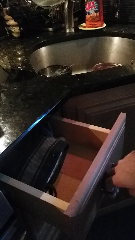}
    \includegraphics[width=\humanwidth]{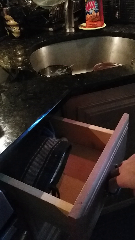}
    
    \includegraphics[width=\humanwidth]{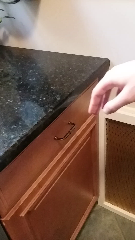}
    \includegraphics[width=\humanwidth]{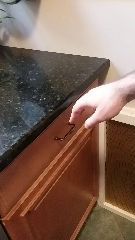}
    \includegraphics[width=\humanwidth]{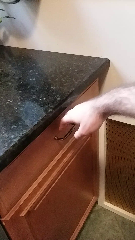}
    \includegraphics[width=\humanwidth]{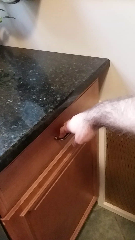}
    \includegraphics[width=\humanwidth]{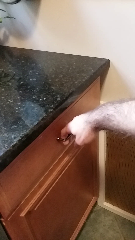}
    \includegraphics[width=\humanwidth]{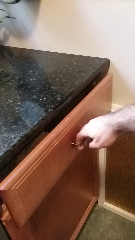}
    \includegraphics[width=\humanwidth]{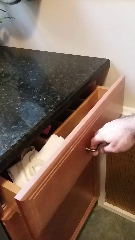}
    \includegraphics[width=\humanwidth]{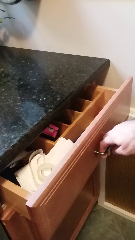}
    
    \includegraphics[width=\humanwidth]{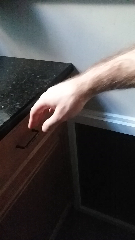}
    \includegraphics[width=\humanwidth]{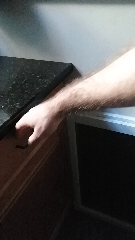}
    \includegraphics[width=\humanwidth]{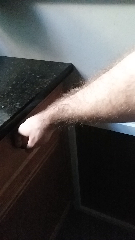}
    \includegraphics[width=\humanwidth]{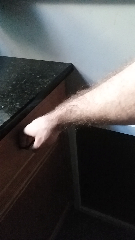}
    \includegraphics[width=\humanwidth]{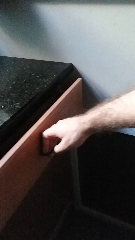}
    \includegraphics[width=\humanwidth]{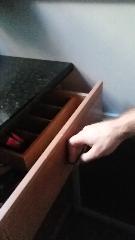}
    \includegraphics[width=\humanwidth]{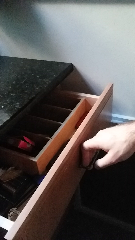}
    \includegraphics[width=\humanwidth]{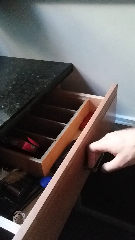}

    \caption{Example images from the human drawer opening observation data used by RLV.}
    \label{fig:human_drawer_dataset_appendix}
\end{figure}
\begin{figure}
\centering
\newcommand{\postwidth}{0.11\linewidth}
    \includegraphics[width=\postwidth]{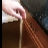}
    \includegraphics[width=\postwidth]{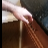}
    \includegraphics[width=\postwidth]{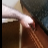}
    \includegraphics[width=\postwidth]{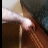}
    \includegraphics[width=\postwidth]{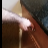}
    \includegraphics[width=\postwidth]{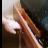}
    \includegraphics[width=\postwidth]{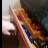}
    \includegraphics[width=\postwidth]{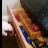}
    
    \includegraphics[width=\postwidth]{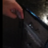}
    \includegraphics[width=\postwidth]{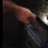}
    \includegraphics[width=\postwidth]{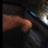}
    \includegraphics[width=\postwidth]{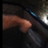}
    \includegraphics[width=\postwidth]{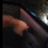}
    \includegraphics[width=\postwidth]{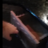}
    \includegraphics[width=\postwidth]{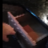}
    \includegraphics[width=\postwidth]{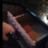}
    
    \includegraphics[width=\postwidth]{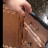}
    \includegraphics[width=\postwidth]{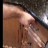}
    \includegraphics[width=\postwidth]{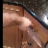}
    \includegraphics[width=\postwidth]{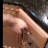}
    \includegraphics[width=\postwidth]{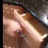}
    \includegraphics[width=\postwidth]{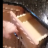}
    \includegraphics[width=\postwidth]{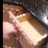}
    \includegraphics[width=\postwidth]{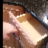}

    \caption{Example post-processed images from the human drawer opening observation data used by RLV.}
    \label{fig:human_drawer_dataset_post_appendix}
\end{figure}

We post-processed the collected videos by transforming them to align the goal square in the pushing data and the drawer handle in the drawer opening data.  The resulting images are then cropped to 48x48 pixels.
Example sequences from the human pushing dataset are shown in Figure~\ref{fig:human_dataset_appendix} and example post-processed images from the human pushing dataset are shown in Figure~\ref{fig:human_dataset_post_appendix}.
Example sequences from the human drawer opening dataset are shown in Figure~\ref{fig:human_drawer_dataset_appendix} and example post-processed images from the human drawer opening dataset are shown in Figure~\ref{fig:human_drawer_dataset_post_appendix}.
The full dataset is available for download on the supplementary website: \url{https://sites.google.com/view/rl-with-videos}.



\end{document}